\documentclass[oneside,english,times,twocolumn,final,authoryear]{elsarticle}
\usepackage[T1]{fontenc}
\usepackage[latin9]{inputenc}
\usepackage{geometry}
\usepackage{color}
\usepackage{babel}
\usepackage{array}
\usepackage{rotating}
\usepackage{float}
\usepackage{rotfloat}
\usepackage{multirow}
\usepackage{algorithm2e}
\usepackage{amsmath}
\usepackage{graphicx}
\PassOptionsToPackage{normalem}{ulem}
\usepackage{ulem}
\usepackage[unicode=true,
 bookmarks=true,bookmarksnumbered=false,bookmarksopen=false,
 breaklinks=true,pdfborder={0 0 1},backref=false,colorlinks=true]
 {hyperref}
\hypersetup{pdftitle={A System For Computational 3D Reconstruction Of Comminuted Bone Fractures},
 pdfauthor={Pengcheng Liu, Nathan Hewitt, Waseem Shadid, Andrew Willis},
 pdfsubject={Medical Image Analysis},
 pdfkeywords={3D puzzle solving; computational 3D reconstruction; bone fracture segmentation; tibial plafond fracture}}

\makeatletter

\providecommand{\tabularnewline}{\\}
\floatstyle{ruled}
\newfloat{algorithm}{tbp}{loa}
\providecommand{\algorithmname}{Algorithm}
\floatname{algorithm}{\protect\algorithmname}

\usepackage{medima}
\usepackage{framed,multirow}
\definecolor{newcolor}{rgb}{.8,.349,.1}

\journal{Computerized Medical Imaging and Graphics}
\jnlurl{www.elsevier.com/locate/compmedimag}

\@ifundefined{showcaptionsetup}{}{%
 \PassOptionsToPackage{caption=false}{subfig}}
\usepackage{subfig}
\makeatother

\begin{document}
\author[uncc]{Pengcheng \snm{Liu}}
\ead{zenonlpc@gmail.com} 
\author[uncc]{Nathan \snm{Hewitt}}
\ead{nhewitt@uncc.edu} 
\author[uncc,leadtools]{Waseem \snm{Shadid}}
\ead{wshadid78@gmail.com} 
\author[uncc]{Andrew \snm{Willis}\corref{cor1}} 
\ead{arwillis@uncc.edu} 
\cortext[cor1]{Corresponding author}
\address[uncc]{University of North Carolina at Charlotte, 9201 University City Blvd., Charlotte, NC 28223-0001, USA}
\address[leadtools]{LEAD Technologies, Inc., 1927 South Tryon Street, Suite 200, Charlotte, NC 28203, USA} 
\begin{frontmatter}

\title{A System for 3D Reconstruction Of Comminuted Tibial Plafond Bone
Fractures}
\begin{abstract}
High energy impacts at joint locations often generate highly fragmented,
or comminuted, bone fractures. Current approaches for treatment require
physicians to decide how to  classify the fracture within a hierarchy
fracture severity categories. Each category then provides a best-practice
treatment scenario to obtain the best possible prognosis for the patient.
This article identifies shortcomings associated with \textit{qualitative}\emph{-only}
evaluation of fracture severity and provides new \emph{quantitative
}metrics that serve to address these shortcomings. We propose a system
to semi-automatically extract quantitative metrics that are major
indicators of fracture severity. These include: (i) fracture surface
area, i.e., how much surface area was generated when the bone broke
apart, and (ii) dispersion, i.e., how far the fragments have rotated
and translated from their original anatomic positions. This article
describes new computational tools to extract these metrics by computationally
reconstructing 3D bone anatomy from CT images with a focus on tibial
plafond fracture cases where difficult qualitative fracture severity
cases are more prevalent. Reconstruction is accomplished within a
single system that integrates several novel algorithms that identify,
extract and piece-together fractured fragments in a virtual environment.
Doing so provides objective \textit{quantitative} measures for these
fracture severity indicators. The availability of such measures provides
new tools for fracture severity assessment which may lead to improved
fracture treatment. This paper describes the system, the underlying
algorithms and the metrics of the reconstruction results by quantitatively
analyzing six clinical tibial plafond fracture cases.
\end{abstract}
\end{frontmatter}

\section{Introduction}

Accurate reconstruction of a patient's original bone anatomy is the
desired outcome for surgical treatment of a bone fracture. Treatment
goals include achieving expeditious reconstruction and avoiding Post-Traumatic
OsteoArthritis (PTOA). When there is involvement of an articulating
joint such as the hip, knee, or ankle, accurate reconstruction of
the bone joint surface is critical to avoid PTOA. This task can be
quite challenging when dealing with highly comminuted fractures. This
is due to the fact that often dozens of individual fragments are involved
and they are sometimes displaced significantly from their original
anatomic position and scattered in a complex geometric pattern.

This article describes a system for 3D medical image and surface analysis
that enables vital new orthopaedic research to improve treatment for
traumatic limb bone fractures. An emphasis is placed on the tibial
plafond fracture as it is both difficult to treat and can exhibit
wide disparity in subjective fracture severity evaluation. This article
details a system constructed as a novel integration of algorithms
that jointly enable virtual 3D reconstruction of a comminuted bone
fracture from a 3D CT image of the fractured limb. While the reconstruction
result could be used for pre-operative planning, this article describes
the application of this system for the purpose of extracting quantitative
data for fracture severity classification.
\begin{figure*}
\noindent \centering{}\includegraphics[height=2.4in]{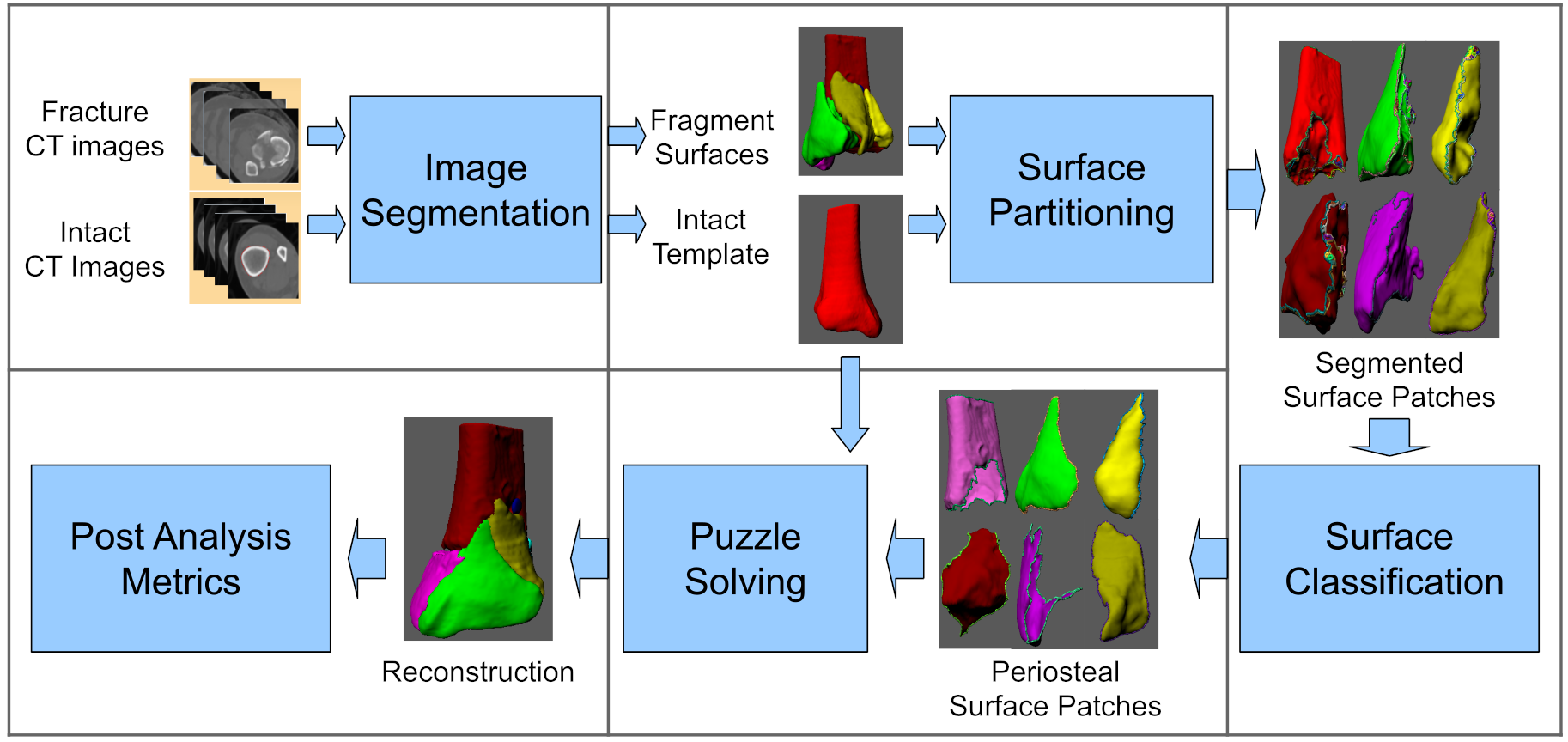}\caption[Overview of the system]{\label{fig:A-brief-overview-of-system-1}A brief overview of a system
proposed in this paper. This system takes as input a 3D CT image of
the fractured limb and a 3D CT image of the undamaged (intact) limb,
and provides as output a virtual reconstruction of bone fragments
which estimates the anatomy of the patient's original bone. Each block
denotes a step and the associated algorithm name is shown inside the
block (in blue). Images show how the data has been changed in each
step of the 5-step reconstruction process.}
\end{figure*}

Accurately classifying the severity of highly comminuted bone fractures
can be challenging for orthopedic physicians and surgeons. Research
in \citep{marsh2002Articular,Beardsley2005,Anderson2008} states that
accurate determination of the initial fracture severity as afforded
by fracture severity classification is the single most important prognostic
determinant of long-term joint health subsequent to trauma. Due to
the importance of fracture severity classification, many researchers
\citep{CPclassify2002,RJclassification1994,ThadDisplay2007,AndersonDD2008,Beardsley2005}
have investigated this problem where their common goal is to define
methods capable of predicting fracture severity from quantitative
measurements derived from medical image data. Yet, no approach to
date provides a holistic solution to this difficult problem as an
integrated system.

Our system accomplishes this task as a sequence of three steps:
\begin{enumerate}
\item Fragment surfaces are extracted from CT images, 
\item Each fragment surface is further decomposed into anatomically meaningful
sub-regions,
\item Fragments are pieced back together in a virtual space with a puzzle-solving
algorithm.
\end{enumerate}
Quantitative data is extracted for fracture severity assessment is
recovered as a by-product of the puzzle-solving solution. The system
enables extraction of previously unavailable quantitative information
from fractures cases in terms of the bone fragment data to assist
physicians in determining the clinical fracture severity classification
of comminuted bone fractures.

Results focus on application of this technology to tibial plafond
fractures. These fractures typically occur as a result of high-energy
trauma such as a ballistic penetration, vehicular accident, or falls
from a height. This article focuses on tibial plafond fracture cases
for the following reasons: 
\begin{enumerate}
\item The complex characteristics of this kind of fracture can often create
difficulties for physicians in making accurate and reliable fracture
severity assessments,
\item Tibia fractures often involve the ankle joint which is typically difficult
to treat, 
\item The quality of reconstruction is a critical factor for good prognosis, 
\item PTOA is directly related to the accuracy of reconstruction.
\end{enumerate}
Hence, this sub-domain of fractures represents one of the most promising
applications of the research described herein.

\section{Related Work}

Computational puzzle solving in 3D seeks to use computer algorithms
to facilitate reconstructing 3D broken objects from the geometry of
their fragments. In general, puzzle-solving approaches fall into three
categories: (1) boundary matching, i.e., algorithms that match together
fragments by comparing their boundaries, (2) template matching, i.e.,
algorithms that match fragments into an \emph{a-priori} known template
that is used as a reference shape for the broken fragments and (3)
manual reconstruction approaches. Approaches from the first two categories
require algorithms for curve and surface matching and surface alignment.
For boundary matching, boundary curves from the broken fragments are
matched to piece together the fragments. For template matching, surfaces
on the fragment are matched to the corresponding surfaces on the template
so that fragments can be aligned into the template to accomplish reconstruction.
\begin{figure}
\subfloat[]{\begin{centering}
\includegraphics[height=1in]{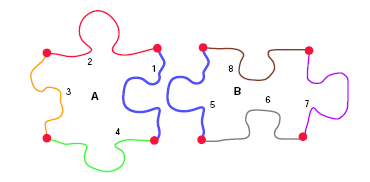}
\par\end{centering}
}\subfloat[]{\begin{centering}
\includegraphics[height=1in]{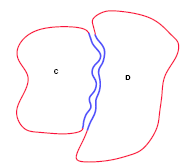}
\par\end{centering}
}

\caption[Difference between reassembling jigsaw puzzles and broken artifacts]{\label{fig:Thejigsawboundary-1}The difference between reassembling
(a) commercial jigsaw puzzles and reconstructing (b) broken artifacts.
Note here that the latter is made much more difficult as corners are
not easily identifiable and may not indicate the beginning or end
of a curve that will uniquely match some other fragment. Worse, any
portion of a boundary curve may match to any other fragment and the
curve itself may match equally well with numerous similar boundaries
from other fragments. (Used with permission from \citet{willis2008reconstructionsurvey}.
)}
\end{figure}

Boundary matching approaches for puzzle solving starts with seminal
work \citep{freeman1964,Wolfson1988} that developed algorithms to
piece together 2D jigsaw puzzles similar to that depicted in figure
\ref{fig:Thejigsawboundary-1}. However, solving 3D bone fracture
``puzzles'' is a significantly more challenging problem. Jigsaw
pieces share similar sizes and have distinctly identifiable shapes
that greatly restrict the collection of potential mating surfaces/curves
on corresponding pieces. Further, for jigsaw puzzles one can assume
that all of the jigsaw pieces are complete and intact, and no pieces
are missing. Puzzle-solving 3D fractures includes low-resolution sensed
data (segmented CT), deformable fragments (bone tissue), potentially
nondescript fracture surfaces (similar to oblique planes) and due
to a high-energy impact there may be small, missing or unusable fracture
pieces.

Fracture reconstruction from boundary data matches boundaries and
surfaces generated due to the fracture event. Some work on bone fragment
reconstruction uses simulated data by simulating fractures via computer-generated
fractures or by using a drop-tower to break bone surrogate material.
Examples include \citep{Kronman2013} which simulates a fracture by
slicing the intact CT scan of a healthy bone. This allows quick experimentation
on a variety of test cases. However, these simulations do not capture
the complex physiological or biomechanical phenomena that give rise
to these fractures which tend to create fractures that may never been
seen in clinical practice. Further, the approach analyzes the geometry
of the fracture surface generated by separating bone fragments and
assumes these surfaces will accurately fit together using some fracture
surface matching metric when in reality bone fragment fracture surfaces
may not geometrically match well. In \citep{Andrew2007Recon} tibial
plafond fractures are simulated using a drop tower to simulate high
axial load to the ankle joint to fracture surrogate bone constructed
from high-density polyether urethane foam. Resulting fragments were
subsequently scanned in a laboratory setting an fracture surfaces
were matched to puzzle-solve the fracture. Here, the measured data
and extracted surfaces exhibit little noise which makes reconstruction
much easier. Work in \citep{Fuernstahl2012} proposes an approach
for reconstructing proximal humerous joint by alignment of fracture
surfaces. A concern here is the geometric accuracy of extracted fracture
surface segmentation; especially in low-contrast trebecular bone tissue
regions. We opt to align outer bone surfaces only avoiding the potentially
inaccurate measurements of the fracture surface regions.

A second category of approaches for bone fragment reconstruction focuses
on aligning fragments into a template of the intact bone. Alignment
is often achieved through applications of the Iterative Closest Point
(ICP) algorithm. This has the distinct disadvantage that, especially
in cases where fragments have undergone large displacements or rotations,
the ICP algorithm can converge on local minima, giving inaccurate
results. In \citep{Chowdhury2009}, work is done on reconstructing
facial fractures. As the geometry here is more unique than the tibia,
the assumption can be made that each bone surface will have only one
match. Because of the differences between comminuted tibia fractures
and craniofacial fractures, the effectiveness of ICP in this context
translates poorly to our application. The work done in \citep{Albrecht2012}
demonstrates that, with modification, the ICP algorithm can improve
upon current reconstruction standards. While a deliberate effort is
made to lessen the likelihood of an erroneous convergence of the algorithm,
it is still possible and is thus an undesirable algorithm. Other work
in \citep{Okada2009} reconstructs fragments of the proximal femur
for fracture cases involving up to 4 fragments by matching fragment
outer surfaces and their boundaries. While multiple fragment solutions
are described, there are examples of a multi-fragment solution and
no computational framework to cope with the large number of potential
matches between fragments, fragment boundaries and their unknown potential
locations. \citep{Moghari2008} defines an Euclidean invariant coordinate
system within which to fit algebraic, i.e., implicit polynomial, surface
representations to bone fragments surfaces and uses the coefficients
of the polynomial models to identify matches between fragments and
their corresponding atlas locations.

Several bone fragment reconstruction methods have been documented
that are dependent on manual input. In \citep{Scheuering2001}, a
system is proposed that simulates volumetric collision detection in
a virtual 3D environment and an optimization process for repositioning
the bone fragments. In this application, users manually place the
fragments close together, then refine the alignment via computer optimization.
This method is made more interactive by the work in \citep{harders2007},
which takes a similar approach of user-guided reconstruction with
collision detection while adding haptic feedback to the system. This
more advanced human-computer interaction increases the intuitiveness
of the system for users. Work in \citep{Cimerman2007,Kovler2015}
proposes user-guided reconstruction with collision detection while
adding surgery simulation. These reconstruction approaches all focus
on manual, i.e., interactive alignment, whose applications include
pre-operative planning, virtual surgical procedures or surgical training.
In contrast, this work targets objective extraction of fracture metrics
for severity analysis. As such, we seek to minimize interaction which
can bias computed results.

This work represents a significant advancement in fracture reconstruction
state-of-the-art in several important ways:
\begin{enumerate}
\item it demonstrates reconstruction solutions for clinical data which include
the most comminuted, i.e., complex, fracture cases in the literature
to date (10 fragments).
\item it incorporates computational acceleration methods required by complex
cases which significantly reduce computational cost.
\item it exhibits smaller average reconstruction alignment errors than observed
in competing solutions \citep{Okada2009,Moghari2008,Fuernstahl2012}.
\end{enumerate}
Further, the end-purpose of the system as a means to facilitate fracture
severity assessment is novel relative to other implementations which
focus on surgical planning.

\section{Methods}

\begin{algorithm*}
\SetAlgoLined
\LinesNumbered

\textbf{Input:} Two 3D CT images: (1) a fractured limb CT, $\mathbf{I}_{f}(x,y,z)$,
and (2) an intact (contralateral) limb CT$,\mathbf{I}_{i}(x,y,z)$.

\textbf{Output:} The reduced fracture poses, $\mathbf{T}_{k}$, and
fracture severity analysis data, ${\cal A}_{k}$.\vspace{1mm}

\hrule

\vspace{1mm}
1: ${\cal S}^{T}={\cal F}\left\{ \mathbf{I}_{i}\right\} $: The template
bone surface ${\cal S}^{T}$ is extracted from the intact CT image
$\mathbf{I}_{i}$. ${\cal {O}}(N^{3})$

2: ${\cal D}_{j}^{T}={\cal E}\left\{ {\cal S}^{T}\right\} $: Surface
descriptors, ${\cal D}_{j}^{T}$, are extracted from the template
bone surface, ${\cal S}^{T}$.

3: ${\cal S}^{k}={\cal F}\left\{ \mathbf{I}_{f}\right\} $: The $k^{th}$
bone fragment surface ${\cal S}^{k}$ is extracted from the fracture
CT image $\mathbf{I}_{f}$.

4: ${\cal S}^{k,o}={\cal C}\left\{ {\cal F}\left\{ {\cal S}^{k}\right\} \right\} $:
Extract the $k^{th}$ fragment outer/cortical surface, ${\cal S}^{k,o}$,
by partitioning and classifying the fragment surface, ${\cal S}^{k}$.$^{*}$

5: ${\cal D}^{k,o}={\cal E}\left\{ {\cal S}_{o}^{k}\right\} $: Extract
a set of surface descriptors, ${\cal D}^{k,o}$, from the $k^{th}$
fragment outer/cortical surface, ${\cal S}^{k,o}$.

6: $\mathbf{T}_{k}={\cal P}\left\{ {\cal D}^{k,o},{\cal D}^{T}\right\} $:
Estimate the $k^{th}$ fragment's pose, $\mathbf{T}_{k}$, from matched
template/fragment feature pairs, $\left\{ {\cal D}_{i}^{k,o},{\cal D}_{j}^{T}\right\} $.$^{*}$

7: ${\cal A}_{k}={\cal L}\left\{ \mathbf{T}_{k}\right\} $: Fracture
severity analysis data, ${\cal A}_{k}$, is extracted from the reduced
fracture solution and surface data, $\left\{ \mathbf{T}_{k},{\cal S}^{k,o}\right\} $.

\caption{\label{alg:Bone-puzzle-solving}Reduction of bone fragment $k$ in
7 algorithmic steps. Processing requires (5) algorithms: (1) an CT
image bone segmentation function, ${\cal F}\left\{ \mathbf{I}\right\} $,
(2) a fragment surface segmentation function, ${\cal F}\left\{ {\cal S}\right\} $,
(3) a fragment surface classifier, ${\cal C}\left\{ {\cal S}\right\} $,
(4) a surface feature descriptor extractor, ${\cal E}\left\{ {\cal S}\right\} $
and (5) a puzzle solver, ${\cal P}\left\{ {\cal D}^{k,o},{\cal D}^{T}\right\} $.
Lines 3-7 are repeated until all fracture fragments are aligned. Steps
ending with an asterisk $^{*}$ above have facilities to assist the
solution via manual interaction (see \S~\ref{subsec:System-Interface-and}
for details).}
\end{algorithm*}
Figure \ref{fig:A-brief-overview-of-system-1} depicts the proposed
approach for fracture reconstruction as a sequence of five steps.
Implementation requires five application-specific algorithms that
process 3D image and surface data. These five algorithms are: 
\begin{enumerate}
\item 3D CT image segmentation (see \S\ref{subsec:Segmenting-Fracture-CT})
\item 3D surface partitioning (see \S\ref{subsec:Partitioning-Subsurfaces})
\item Appearance-based 3D surface classification (see \S\ref{subsec:Appearance-Based-3D-Surface})
\item 3D puzzle-solving (see \S\ref{subsec:Computational-3D-Puzzle})
\item Quantitative fracture analysis (see \S\ref{subsec:Post-Reconstruction-Analysis})
\end{enumerate}
A detailed list of processing steps, including the intact template
processing is described in Algorithm \ref{alg:Bone-puzzle-solving}.
These steps also specify the sequence of algorithms through which
the data flows as each algorithm requires results computed from the
previous one.

\subsection{\label{subsec:Computational-Complexity-Analysi}Computational Complexity
Analysis}

We provide a computational complexity analysis of our entire system
by assessing the complexity of each of the steps required to compute
the reconstruction, i.e., steps 1-4 above. The segmentation algorithm
of step (1) has a computational complexity ${\cal {O}}(M+N\log(N))$
\citet{Felkel2001} where $N$ denotes the total number of voxels
in the medical image and $M$ denotes the number of edges between
voxels as defined by the adjacency scheme. The algorithm we use, \citep{SHADID201814,Shadid2013},
implements a 6-adjacency scheme which $M\simeq N$ and the resulting
algorithm have complexity ${\cal {O}}(M+N\log(N))$. The surface partitioning
algorithm of step (2) uses the graph of the surface mesh to partition
a closed surface into surface patches. Surface partitions are defined
using a minimum spanning tree of the surfaces which has computational
complexity ${\cal {O}}(E\log(V))$ for a mesh surface having $E$
edges and $V$ vertices. The classification algorithm of step (3)
requires ${\cal {O}}(N)$ calculations to complete. The 3D puzzle
solving algorithm of step (4) consists of two parts: (i) a surface
correspondence search and (ii) candidate match evaluation. The surface
correspondence search problem is solved via brute-force pairwise evaluation
of $N^{2}$ surface correspondences. Each correspondence is assigned
a similarity score by matching spin image surface features, an algorithm
that has linear computational complexity in the spin image descriptor
dimension, i.e., ${\cal {O}}(N_{overlap})$, where $N_{overlap}$
denotes the number of overlapping pixels in a spin image pair match
(see \S~\ref{subsec:Feature-Extraction}). A small subset of surface
matches having high similarity scores are processed using the ICP
algorithm to compute the final fracture reconstruction which has computational
complexity ${\cal {O}}(kV_{o}^{2})$ where $V_{o}$ denotes the number
of surface vertices and $k$ denotes the number of iterations required
for ICP to converge. We mention $k$ explicitly to point out that
our method for coarsely aligning surfaces via spin images can significantly
reduce $k$ by starting this nonlinear optimization closer to the
desired minimum of the error functional.

\subsection{\label{subsec:System-Interface-and}System Interface and Interactions
to Assist Reconstruction}

Our system can reconstruct fracture cases completely automatically
but include interfaces to specify algorithm parameters. For some steps,
i.e., steps (4) and (6) of from Algorithm \ref{alg:Bone-puzzle-solving}
an interface has been included to allow manual interaction to steer
the system to a better solution. Step (4) from Algorithm \ref{alg:Bone-puzzle-solving}
requires access to a CT-machine specific training database to classify
fragment surfaces. If this data is not available, an interface is
provided that allows the user to manually classify fragments surfaces
with may take several minutes to specify interactively. Step (6) from
Algorithm \ref{alg:Bone-puzzle-solving} also has an interactive interface
which allows users to manipulate fragments in the final solution.
This interaction is rarely used and typically is applied to re-initialize
fine alignment, e.g., one may slightly rotate a small fragment and
initiate new a global puzzle alignment solution to find a better solution.

\subsection{\label{subsec:Segmenting-Fracture-CT}Segmenting Fracture CT Images}

Reconstruction of a 3D solid from its fragments is a geometric problem
where the geometry of the fragments must be known in order to piece
them back together. For this reason geometric models for each bone
fragment in the 3D CT image must be computed. This is the goal of
the CT image segmentation process. The segmentation algorithm used
assigns each pixel in the CT image $\mathbf{I}(x,y,z)$ to a label,$l$.
The goal of the surface segmentation algorithm is to estimate the
correct label for every pixel and we denote this operation is mathematically
with ${\cal F}\left\{ \mathbf{I}\right\} $. Fragment surfaces may
be extracted from the labeled image by estimating the locations where
the image changes label, or equivalently, solving for the locations
where the label image changes from fragment label, $l=k$, to a different
label value, i.e., ${\cal S}^{k}=\left\{ {\cal F}\left\{ \mathbf{I}_{f}\right\} \,\,|\,\,\left\Vert \nabla{\cal F}\left\{ \mathbf{I}_{f}\right\} \right\Vert \neq0,{\cal F}\left\{ \mathbf{I}_{f}\right\} =k\right\} $.
Geometric segmentation of bone fragments is challenging due to the
similarity in cancellous and background CT tissue intensities making
standard \citep{Eijnatten2018} or manual \citep{Paulano2014} approaches
impractical. This work adopts the approach described in \citep{SHADID201814,Shadid2013}
to segment the image which improves upon prior watershed algorithm
implementations \citep{F.Meyer1994}; especially for bone fragment
segmentation. Bone fragment surfaces are extracted from the segmented
image using the marching cubes algorithm \citep{Lorensen1987MarchingCube}.
There are two major challenges to this segmentation problem:
\begin{figure}
\begin{centering}
\subfloat[]{\begin{centering}
\includegraphics[width=0.2\columnwidth]{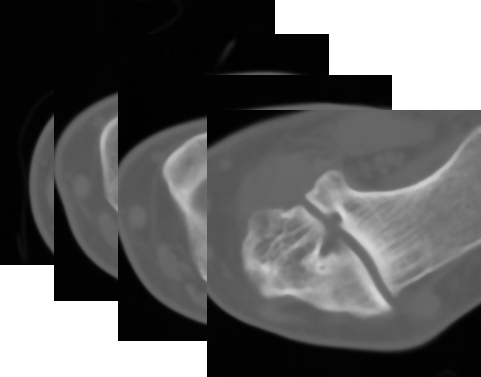}
\par\end{centering}
}\subfloat[]{\begin{centering}
\includegraphics[width=0.2\columnwidth]{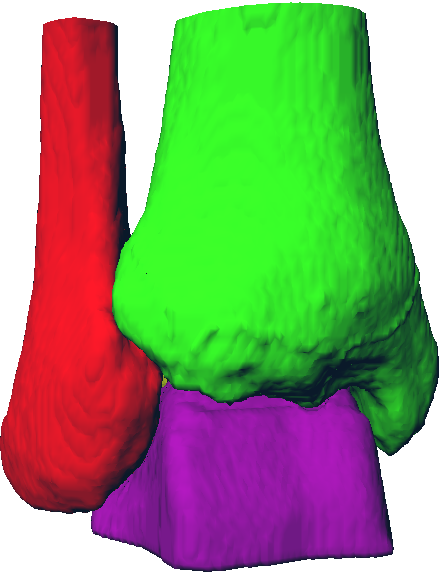}
\par\end{centering}
}\subfloat[]{\begin{centering}
\includegraphics[width=0.2\columnwidth]{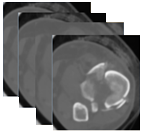}
\par\end{centering}
}\subfloat[]{\centering{}\includegraphics[width=0.2\columnwidth]{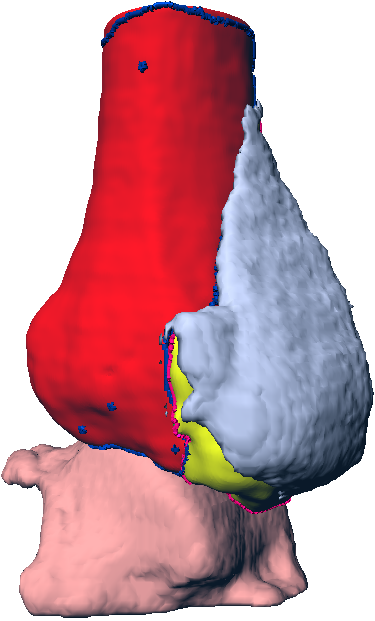}}
\par\end{centering}
\centering{}\caption{(a) shows slices from a 3D CT image of an intact ankle joint. (b)
shows the segmented tibia, fibula, and talus bone surfaces. (c) shows
slices from a 3D CT image of a fractured ankle. (d) shows the segmented
bone fragments surfaces. The segmentation from the 3D CT image was
performed using the proposed segmentation algorithm.\label{fig:1}}
\end{figure}

\begin{enumerate}
\item Bone fragment boundaries are especially difficult to demarcate when
fragments are abutting other fragments.
\item Intensities for some bone tissues, i.e., cancellous tissue, are the
same as that for other soft tissues in the same anatomic region making
them difficult to discriminate.
\end{enumerate}
The bone fragment segmentation algorithm takes as input a CT image
of the limb and two user parameters: (1) the maximum size of a bone
region, and (2) a segmentation sensitivity parameter. The output of
this algorithm is a labeled CT image where each unique label corresponds
to a unique bone fragment.

The segmentation algorithm proceeds by incrementally classifying pixels
from image regions of high-likelihood (cortical tissue) to regions
of low-likelihood. To do so, image pixels are initially classified
into three sets: (1) non-bone, (2) cortical bone, and (3) non-cortical
bone. Connected cortical bone pixel regions are assigned a fragment
label and a customized watershed algorithm \citep{Shadid2013}, referred
to as the Probabilistic Watershed Transform (PWT), merges adjacent
pixels into each fragments region. The merge procedure calculates
the conditional probability of each candidate pixel given the current
segmentation and includes models to specifically address the realization
of layered, i.e., lamellar bone tissue structures in CT images, and
common fracture phenomenon such as fragment splintering which generates
long, thin protrusions of bone tissue that is difficult to otherwise
segment. Surfaces are extracted from the segmented image data with
the marching cubes algorithm \citep{Hansen2005}. Figures (\ref{fig:1}(b,d))
show 3D surfaces segmented from CT images of ankle joints for an intact
(b) and a fractured (d) ankle joint using the proposed algorithm.

\subsection{\label{subsec:Partitioning-Subsurfaces}Partitioning Surfaces}

Puzzle-solving requires shapes to be decomposed into shape parts.
Similarly, for fracture reconstruction, each segmented bone fragment
model must be partitioned into a collection of surface patches to
allow the outer surfaces of each fragment surface to be geometrically
matched to the template surface, ${\cal S}^{T}$. A surface segmentation
algorithm divides fragment surfaces into 3 anatomically distinct groups
which greatly reduces the difficulty of the search problem by limiting
the number of plausible correct surface matches and thereby improving
the reconstruction system performance. The surface mesh partitioning
algorithm divides the $k^{th}$ bone fragment surface, $S^{k}$, into
a collection of surface patches, $\{s_{1},s_{2}\cdots\}$ , such that
$S^{k}=\cup\{s_{1},s_{2}\cdots\}$. Each of the generated surface
patches are intended to consist of surface points from only one of
three anatomical categories. To do so, the system applies a ``ridge
walking'' algorithm \citep{andrew2010ridge} which solves this problem
geometrically using the fact that the anatomical categories of interest
can be discriminated well by dividing the 3D bone fragment surfaces
along 3D contours that traverse high-curvature ridges. 

\subsection{\label{subsec:Appearance-Based-3D-Surface}Appearance-Based 3D Surface
Classification}

\begin{figure}
\begin{centering}
\includegraphics[width=0.8\columnwidth]{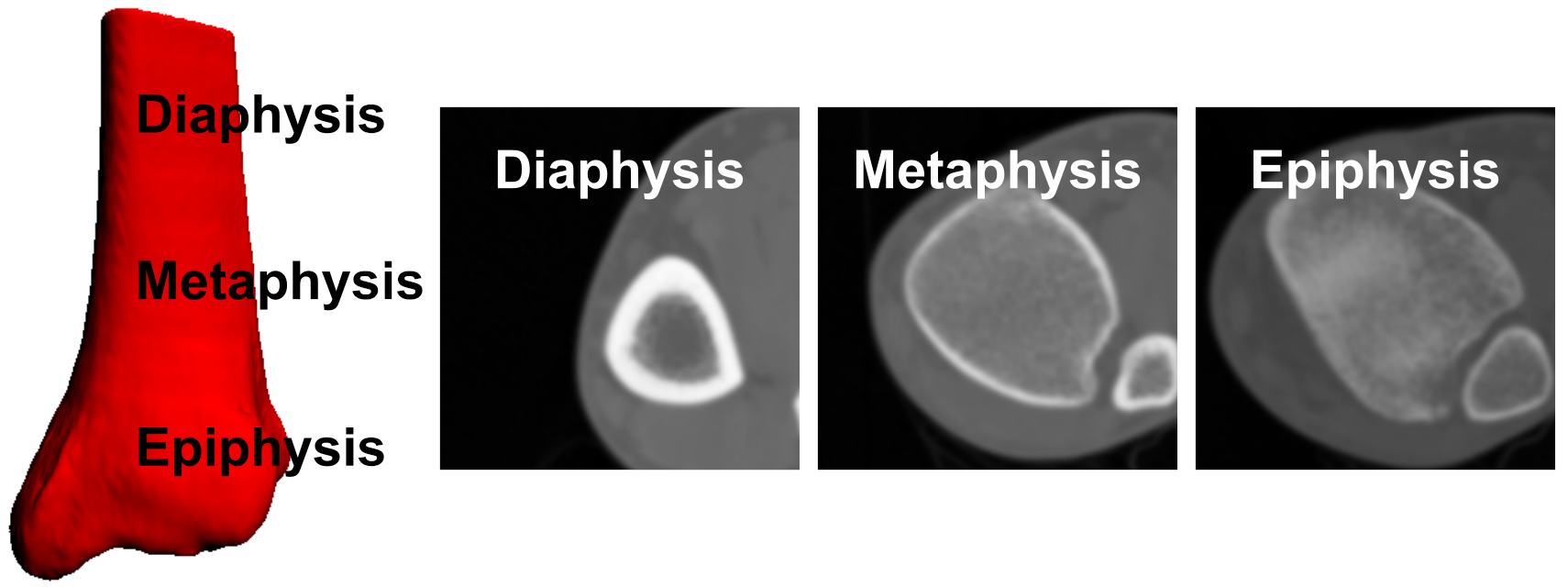}
\par\end{centering}
\caption[CT appearance of the distal tibia anatomy]{\label{fig:CT-appearance-of-1}CT appearance of the distal tibia
anatomy. Relative to the tibia\textquoteright s outer surface, the
CT intensities vary with characteristic profiles along the inward
surface normal for each anatomic region.}
\end{figure}

\begin{figure}
\begin{centering}
\includegraphics[height=1in]{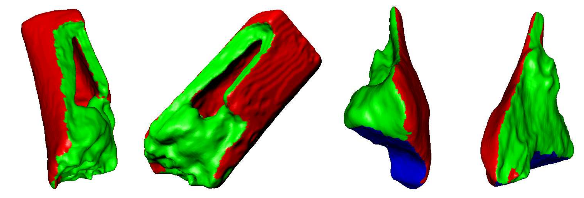}
\par\end{centering}
\caption[Surface mesh classification results]{\label{fig:Surface-mesh-classification-1}Two views of the surface
mesh classification results for two fragments. Red denotes periosteal
surface, green denotes fracture surface, and blue denotes articular
surface.}
\end{figure}

A surface classification algorithm must identify the outer, i.e.,
cortical, surface(s) of each bone fragment from the collection surface
patches created by the surface partitioning algorithm. This is a classification
problem that is solved using by a classification algorithm, ${\cal C}\left\{ s_{1},s_{2}\cdots\right\} $
that extracts the periosteal, i.e., outer, surface, ${\cal S}^{k,o}$,
from the set of partitioned surfaces, $\left\{ s_{1},s_{2}\cdots\right\} $
from the $k^{th}$ fragment for template-based reconstruction. Our
novel classification approach uses CT bone tissue intensity variations
in the vicinity the fragment surface to detect the unknown anatomic
label of the surface. This is made possible by the fact that cortical,
trebecular, cancellous and subchondral bone tissues have distinct
intensity ranges and relative thicknesses in the anatomic regions
of interest. Unfortunately, intensity variations by machine, patient
gender and age \citep{Looker_2009,Felson1993} require the user to
specify training data from within the CT image for reliable output.
Figure \ref{fig:CT-appearance-of-1} shows image data from the three
anatomic bone regions that must be marked interactively by a user:
(1) the diaphysis, made of solid dense cortical bone, (2) the metaphysis,
made of a cortical shell and an interior, porous, cancellous bone,
and (3) the epiphysis, made of dense subchondral bone. From this data,
the automated classification algorithm \citep{liu2012} assigns the
partitioned fragment surface patches to one of three semantic labels:
(1) \textquotedblleft fracture surfaces\textquotedblright{} (surfaces
generated when the bone broke apart), (2) \textquotedblleft periosteal
surfaces\textquotedblright{} (surfaces that were part of the outer
bone surface), and (3) \textquotedblleft articular surfaces\textquotedblright{}
(surfaces that facilitate the articulation of joints). Classification
of these surface patches is useful for both bone reconstruction and
severity analysis. For reconstruction, the periosteal surfaces for
geometric alignment for fragments template-based reconstruction (the
approach used in this paper), (2) fracture surfaces allow calculation
of surface area produced by a fracture event, which is a valuable
objective severity metric, and (3) articular surfaces are particularly
medically important in reconstruction. From this data a classifier
is constructed and applied to automatically classify the extracted
surface patches to these three anatomic labels. A final step merges
adjacent periosteal surface patches having the same anatomic label
generating the periosteal, i.e., outer, surface, ${\cal S}^{k,o}$.
Figure \ref{fig:Surface-mesh-classification-1} shows classification
results for two bone fragments using this approach.

\subsection{Computational 3D Puzzle Solving\label{subsec:Computational-3D-Puzzle}}

The 3D puzzle-solving algorithm takes as input the $k^{th}$ fragment
outer surface patch, ${\cal S}^{k,o}$, and estimates the transformation,
$\mathbf{T}_{k}$, that rotates and translates this fragment to it's
original anatomic position within the intact template. A complete
virtual reconstruction of the unbroken bone from the bone fragments
is achieved by transforming the geometry of each fragment using it's
associated transformation. The intact contra-lateral bone, as represented
in the intact CT image, is taken as a reasonable approximation of
the unbroken bone after mirroring the geometry across the human bilateral
plane of symmetry\sout{.}

\subsubsection{Generic template matching as a puzzle solving approach\label{sec:Generic-template-matching}}

Puzzle solving algorithms solve a difficult computational search problem
whose goal is to estimate the unknown geometric correspondences between
puzzle pieces. This is typically accomplished in two steps: (1) hypothesize
geometric matches and (2) test hypothesized matches by evaluating
a test statistic; typically the minimized pairwise surface alignment
error using the ICP algorithm \citep{Besl1992,Chena,Koklim2004ICP,Zhang:1994:IPM:195967.195970}).
Each hypothesis has the form: ``Fragment surface, ${\cal S}^{k,o}$,
corresponds template surface, ${\cal S}^{T}$, at matching surface
points $\left\{ \mathbf{p}_{k},\mathbf{p}_{T}\right\} $ respectively.''
For the $k^{th}$ puzzle fragments the puzzle-solved fragment pose,
$\mathbf{T}_{k}$, is taken as the pairwise hypothesis which, after
alignment, has the smallest alignment error.

Use of the ICP algorithm to puzzle solve the fracture is both an unreliable
and computationally prohibitive option. Reliability is an issue due
to the tendency of the ICP algorithm to converge to the closest local
minima. Hence, unless the fragments are close to their final anatomic
pose in the puzzle solution, it is unlikely their final alignment
will be correct. Computational cost is an issue since many fragment
surfaces include thousands of samples and finding corresponding surface
points is a ${\cal {O}}(N^{2})$ search problem and evaluation of
each pairwise correspondence requires a iterative ICP solution which
has computational complexity ${\cal {O}}(N^{2})$. Hence the total
computational cost of ICP-based reconstruction for one fragment is
${\cal O}(N^{4})$ where $N$ is the number of surface points. For
$m$ fragments the complexity increases to ${\cal {O}}(mN^{4})$ ($m\ll N)$
\citep{Arthur2006}. For our experiments, $N^{4}\sim10^{20}$ which
imposes computational costs high enough to make brute force solutions
computationally prohibitive for even very small puzzles.

\subsubsection{\label{sec:The-feature-based-puzzle}Surface Descriptor-Based Puzzle
Solving}

\begin{figure*}
\begin{centering}
\includegraphics[height=2in]{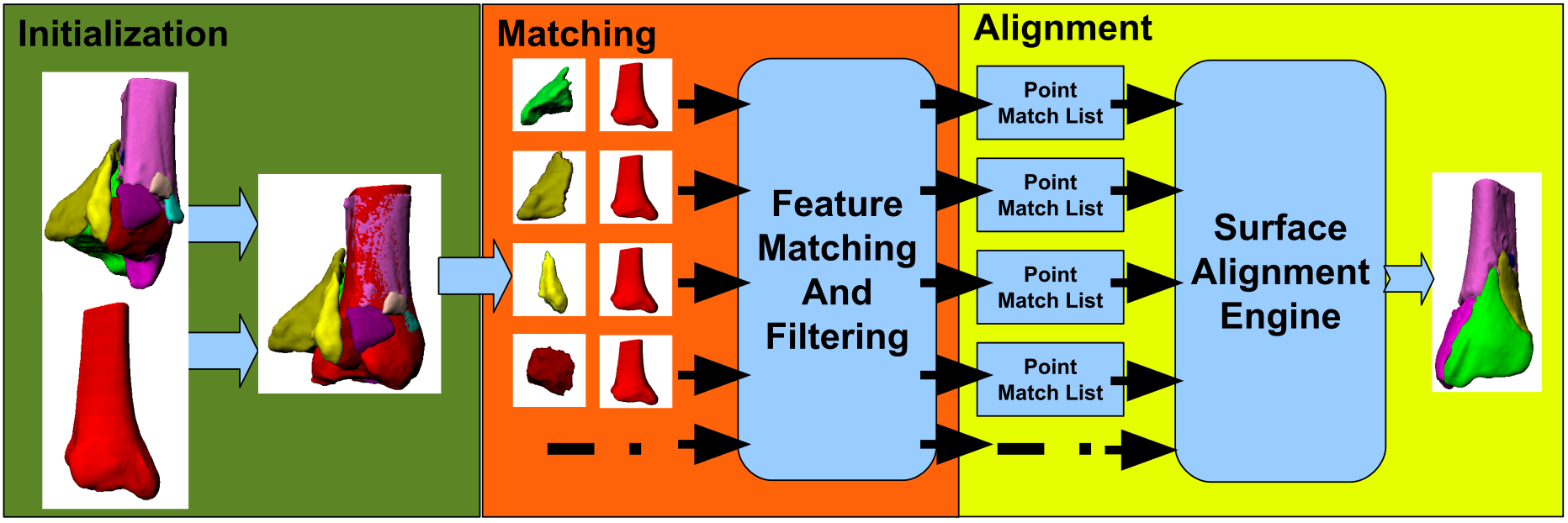}
\par\end{centering}
\caption[Puzzle solving process]{\label{fig:Puzzle-soving-process:}Puzzle solving process: initialization,
matching and alignment. }
\end{figure*}

An approach for bone fragment puzzle solving is proposed to make the
puzzle solving algorithm computational cost tractable. Descriptor-based
puzzle solving approaches extracts descriptors from surfaces and uses
these descriptors to hypothesize matches and geometrically align the
pieces and use the average alignment error as a statistic to evaluate
the likelihood that the hypothesized match is correct. The approach
to puzzle-solve the $k^{th}$ fragment, $S^{k,o}$, with respect to
intact template, $S^{T}$, consists of five steps: 
\begin{enumerate}
\item \textbf{Descriptor Extraction}: This step applies a surface descriptor
extractor algorithm, ${\cal E}\left\{ {\cal S}\right\} $, to puzzle
surfaces, ${\cal S}$, to produce a set of Euclidean-invariant surface
descriptors. Let ${\cal D}^{T}={\cal E}\left\{ {\cal S}^{T}\right\} $
denote descriptors extracted from the template surface and ${\cal D}^{k,o}={\cal E}\left\{ {\cal S}^{T}\right\} $
denote descriptors from the $k^{th}$ fragment surface. (see \ref{subsec:Feature-Extraction})
\item \textbf{Generate Matched Features:} This step takes in as input the
computed descriptors from (1) and outputs a list of matched descriptor
pairs: $L_{initial}$. (see \ref{subsec:Feature-Matching})
\item \textbf{Remove Incorrect Matches:} This step takes $L_{initial}$
as input and removes suspected false matches from the initial list
and outputs a list of candidate descriptor matches in a new list:
$L_{candidate}$. (see \ref{subsec:Filtering-Feature-Matches})
\item \textbf{Test Candidate Matches:} This step takes in the previously
generated list $L_{candidate}$ and outputs a 3D transformation matrix,
$\mathbf{T}_{k}$, for each match which aligns the bone fragment to
the surface of the intact template.(see \ref{subsec:Surface-Data-Matching})
\item \textbf{Select The Best Matches:} This step determines which of the
alignments from the previous step provide the best result and uses
these to provide the final puzzle-solved solution. (see \ref{subsec:Output-The-Solution})
\end{enumerate}
Applications in surgical planning and image guided surgery require
the puzzle solution to be geometrically registered to the coordinate
system of the fractured CT image, $\mathbf{I}_{f}$. To this end,
an initial alignment is performed between intact template surface,
${\cal S}^{T}$, and the anatomically un-perturbed fragment of the
fractured limb which we refer to as the ``base fragment.'' For tibial
plafond fractures the ``base fragment,'' is typically the uppermost
(proximal) bone fragment in the fracture. This initial movement of
the template serves to bring the template surface into the coordinate
system and vicinity of the bone fragment surfaces improving convergence
rates of iterative geometric alignment optimizations. Figure \ref{fig:Puzzle-soving-process:}
shows a graphical overview of this method for puzzle-solving a clinical
bone fracture case.

\subsubsection{\label{subsec:Feature-Extraction}Feature Extraction}

\begin{figure}
\begin{centering}
\subfloat[\label{fig:oriented-surface-p}]{\includegraphics[width=0.22\columnwidth,height=0.1\textheight]{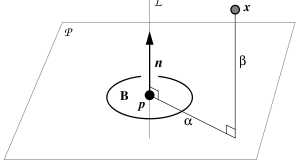}

}\subfloat[\label{fig:point-p-on-surface}]{\includegraphics[width=0.22\columnwidth,height=0.1\textheight]{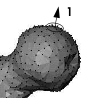}

}\subfloat[\label{fig:plotting-grid}]{\includegraphics[width=0.22\columnwidth,height=0.1\textheight]{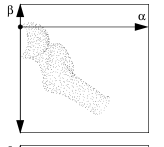}

}\subfloat[\label{fig:spin-image}]{\includegraphics[width=0.22\columnwidth,height=0.1\textheight]{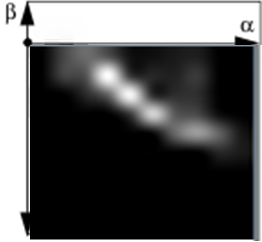}

}
\par\end{centering}
\caption[Example of computing a spin image]{(a) Oriented surface point $\mathbf{p}$ with its normal n and the
tangent plane $p$, one of its neighbor point $\mathbf{x}$ on the
fragment surface and $(\alpha$,$\beta)$ are computed values for
point $\mathbf{x}$ respect to point $\mathbf{p}$ ; (b) shows point
$\mathbf{p}$ on the bone surface and its neighbor points; (c) Plotted
neighbor points on the $(\alpha,\beta)$ grid; (d) Computed spin image
for out of (c) \citet{Johnson:1997:SRM:523428.825373}. }
\end{figure}

While there are many candidate surface descriptors that could be used
to compactly characterize local surface patch geometry, this paper
adopts the spin image representation \citep{Johnson:1997:SRM:523428.825373,Johnson96recognizingobjects}
as its surface descriptor. The rationale for this choice is based
on the low computational cost this representation affords when performing
surface matching. Specifically, spin images converts a nonlinear 3D
surface-pair alignment problem into a low-dimensional 2D (spin) image-pair
comparison problem (see figure \ref{fig:spin-image}). Spin images
of two identical surface patches in arbitrary pose generate produce
identical 2D spin images. As mentioned in \S~\ref{subsec:Computational-Complexity-Analysi},
use of spin images significantly reduces the complexity of comparing
surface shapes by reducing the computational complexity from polynomial
time ${\cal {O}}(kN^{2})$(ICP) to linear ${\cal O}(N_{overlap})$
for spin image pair having $N_{overlap}$ overlapping pixels (2D-image
subtraction). Unfortunately, matching spin images does not solve the
complete alignment problem and requires estimation of a single additional
parameter: rotation about the corresponding surface normals. Hence,
spin images provide a compact, yet imperfect, Euclidean invariant
representation of the shape of a bone fragment surface.

\subsubsection{\label{subsec:Feature-Matching}Generate Matched Descriptors}

Let $L_{initial}$ denote the list of initial hypothesized matches
where each match is specified by an index pair $\left\{ i,j\right\} $
indicating that the $i^{th}$ surface descriptor from the template,
${\cal D}_{i}^{T}$, is hypothesized to correspond to the $j^{th}$
surface descriptor from the $k^{th}$ fragment's outer surface ${\cal D}_{j}^{k,o}$.
Evaluation of the correspondence hypotheses is accomplished by calculating
the value of a similarity function $C_{sp}(P,Q)$ to detect similar
surface matches. To avoid biases in descriptor matching, the proposed
function combines statistical correlation and sample size into this
similarity function.

Our similarity function uses\sout{ }the normalized linear correlation
coefficient, $R_{sp}(P,Q)$, to measure descriptor similarities as
shown in equation (\eqref{eq:matchingsp}) which assigns a score of
1 for a perfect match and scores can range from (-1, 1). Given two
spin images $P$ and $Q$ with $N_{bin}$ bins (number of pixels in
the image ), denote $p_{i}=P(\alpha,\beta)$ as the $i^{th}$ spin
image intensity value from spin image $P$, similarly, denote $q_{i}=I(\alpha,\beta)$
as the $i^{th}$ spin image intensity value from spin image $Q$.
With this notation, the linear correlation coefficient, $R_{sp}(P,Q)$,
is computed as shown in equation \eqref{eq:matchingsp}:

{\footnotesize{}
\begin{equation}
R_{sp}(P,Q)=\frac{N_{bin}\underset{i}{\sum}p_{i}q_{i}-\underset{i}{\sum}p_{i}\underset{i}{\sum}q_{i}}{\sqrt{\left(N_{bin}\underset{i}{\sum}p_{i}^{2}-\left(\underset{i}{\sum}p_{i}\right)^{2}\right)\left(N_{bin}\underset{i}{\sum}q_{i}^{2}-\left(\underset{i}{\sum}q_{i}\right)^{2}\right)}}\label{eq:matchingsp}
\end{equation}
}{\footnotesize\par}

When $R_{sp}$ is close to 1, the images are similar, when $R_{sp}$
is close to -1 the images are different. Unfortunately, this metric
is biased since each spin image may have different $(\alpha,\beta)$
dimensions. Hence good matches may be found when spin images have
similar values over small overlapping $(\alpha,\beta)$ regions. On
the other hand, two spin images with a large area of overlap may be
assigned a lower $R_{sp}$ value but exhibit widespread similarities. 

To address biases $R_{sp}(P,Q)$ the similarity function $C_{sp}(P,Q)$
proposed in \citep{Johnson:1997:SRM:523428.825373,Johnson96recognizingobjects}
was used as shown in equation (\ref{eq:matchingsimisp}). This metric
is well-suited because it considers both the correlation, i.e., match
fidelity, and also the number of matching or overlapping pixels, $N_{overlap}$
, as part of the final matching score. 
\begin{equation}
C_{sp}\left(P,Q\right)=\left(\text{{atanh}}\left(R_{sp}\left(P,Q\right)\right)\right)^{2}-\lambda\left(\frac{1}{N_{overlap}-3}\right)\label{eq:matchingsimisp}
\end{equation}

The resulting similarity function weights the correlation $R_{sp}$
against the variance intrinsic to the correlation coefficient which
increases as the amount of overlap in the two spin images increases.
This is accomplished via a change of variables commonly referred to
as \textquotedbl{}Fisher's z-transformation\textquotedbl{} \citep{Fisher:z-transformation}
which uses the hyperbolic arctangent function to transform the distribution
of the correlation coefficient into an approximately normal distribution
having constant variance $\frac{1}{N_{overlap}-3}$ . This allows
the similarity metric (\ref{eq:matchingsimisp}) to be formed that
strikes a compromise between good correlations (term 1) and sufficient
evidence, i.e., measurements to trust the computed correlation (term
2) to better detect reliable matches. The parameter $\lambda$ is
used to control the relative weight of the expected value of the correlation
coefficient and the variance of this statistic to produce a final
similarity score. \citep{Johnson:1997:SRM:523428.825373,Johnson96recognizingobjects}
mention that $\lambda$ controls the point at which the overlap between
spin images dominates the value of the similarity metric for two spin
images. When the overlap is much larger than $\lambda$, the second
term in equation (\ref{eq:matchingsimisp}) becomes negligible. In
contrast, when the overlap is much less than $\lambda$, the second
term dominates the similarity measure. Therefore, $\lambda$ should
be the expected overlap between spin images. In this paper, $\lambda$
is automatically computed for each surface match by computing average
number of non-black pixels for all the spin images generated from
that fragment and setting $\lambda$ to half of the average value.

\subsubsection{\label{subsec:Filtering-Feature-Matches}Removing Incorrect Matches}

Due to the noise from image data and errors from segmentation and
classification algorithms, the initial list of matches $L_{initial}$
often contains many false hypothesized surface point correspondences.
Since careful analysis of these matches is computationally expensive,
we propose an alternative approach to detect incorrect hypotheses
that significantly improves performance. To do so we consider multi-hypothesis
consistency constraints to efficiently eliminate implausible solutions.

Conceptually, the idea here is that each fragment has a single unique
Euclidean transformation that brings the fragment to it's original
anatomic location, $\mathbf{T}_{k}$. Hence, if multiple hypotheses
from the same fragment are made, those that are correct must have
nearly the same value for $\mathbf{T}_{k}$. Towards this end, evaluated
hypotheses for the $k^{th}$ can be used to cross-validate other hypotheses
from the same fragment. However, rather than performing the computationally
expensive alignment required to compute $\mathbf{T}_{k}$, we use
similarity as provided by spin image surface descriptors to reject
large numbers of incorrect hypotheses.

Let $\left\{ {\cal D}_{i}^{T},{\cal D}_{j}^{k,o}\right\} $ and $\left\{ {\cal D}_{l}^{T},{\cal D}_{m}^{k,o}\right\} $
denote descriptor pairs $\{i,j\}$ and $\{l,m\}$ from the list $L_{initial}$
computed from surface points are $\left\{ \mathbf{p}_{i}^{T},\mathbf{p}_{j}^{k,o}\right\} $
and $\left\{ \mathbf{p}_{l}^{T},\mathbf{p}_{m}^{k,o}\right\} $. If
both hypotheses are correct, then the surface point pairs should be
geometrically consistent, i.e., the distance between $\mathbf{p}_{j}^{k,o}$
and $\mathbf{p}_{m}^{k,o}$ should be equal to the distance between
$\mathbf{p}_{i}^{T}$ and $\mathbf{p}_{l}^{T}$. This constraint can
be directly validated using spin image coordinates in a geometric
consistency test. If the two matches satisfy equation (\eqref{eq:filtermatches}),
we consider them to be geometrically consistent matches which means
both of them may be true matches.

\begin{equation}
\left|{\cal D}_{i}^{T}\left(\mathbf{p}_{j}^{k,o}\right)-{\cal D}_{j}^{k,o}\left(\mathbf{p}_{i}^{T}\right)\right|-\left|{\cal D}_{l}^{T}\left(\mathbf{p}_{m}^{k,o}\right)-{\cal D}_{m}^{k,o}\left(\mathbf{p}_{l}^{T}\right)\right|<D_{gc}\label{eq:filtermatches}
\end{equation}

Where $D_{gc}=2\gamma_{intact}$ where $\gamma_{intact}$ is the resolution
of the intact template, i.e., the average edge length of the edges
from the intact template surface mesh. 

When the initial match list is constructed, list elements are sorted
by decreasing similarity score, $C_{sp}$. Inconsistent elements are
removed by splitting the list $L_{initial}$ into two parts at its
midpoint. The first list contains matches having higher similarity
scores and the second contains matches that having lower similarity
scores. We consider one match from each of the two lists respectively
form pairwise correspondence hypotheses. The geometric consistency
condition is evaluated for the pair of correspondences. If they both
satisfy the consistency condition, we keep both matches. Otherwise,
we keep the match that has a higher similarity score and discard the
other match. After evaluating all matches in both lists, the remaining
matches are placed in the sorted list referred to as $L_{candidate}$. 

\subsubsection{\label{subsec:Surface-Data-Matching}Test Candidate Matches}

The reduced set of hypotheses in the $L_{candidate}$ set are now
evaluated by calculating their 3D surface alignment error. As mentioned
previously, this is a computationally expensive step involving nonlinear
optimization via the ICP algorithm which is further known to suffer
from erroneous solutions when the initial transformation, $\mathbf{T}_{0}$,
for optimization lies far from the true value, $\mathbf{T}_{k}$.
Our surface descriptor matches serve to significantly reduce computation
here by performing a coarse initial alignment which provides a good
initial guess for 5 of the 6 unknown transformation parameters, specifically
the (3) translation parameters and (2) of the three unknown rotation
parameters. For a given match, the alignment proceeds in two steps: 
\begin{enumerate}
\item Coarse Surface Alignment: For each likely descriptor match hypothesis
$\left\{ {\cal D}_{i}^{T},{\cal D}_{j}^{k,o}\right\} $ an initial
transform, $\mathbf{T}_{0}$, is calculated which translates the fragment
surface,${\cal S}^{k,o}$, to make it's surface point, $\mathbf{p}_{j}^{k,o}$,
correspond with the template's surface point $\mathbf{p}_{i}^{T}$
and simultaneously rotates the fragment surface to make the fragment
and template surface normals also correspond at these points.
\item Refined Surface Alignment: Using the initial transform $\mathbf{T}_{0}$,
we run the ICP algorithm to compute final 3D alignment error.
\end{enumerate}
In this way hypotheses can be used that may not be exactly correct
and in cases where the hypothesis is ``close,'' the ICP algorithm
is still likely to converge to the correct alignment.

\subsubsection{\label{subsec:Output-The-Solution}Selecting The Best Matches}

\begin{figure}
\begin{centering}
\includegraphics[width=1\columnwidth,height=0.2\textheight,keepaspectratio]{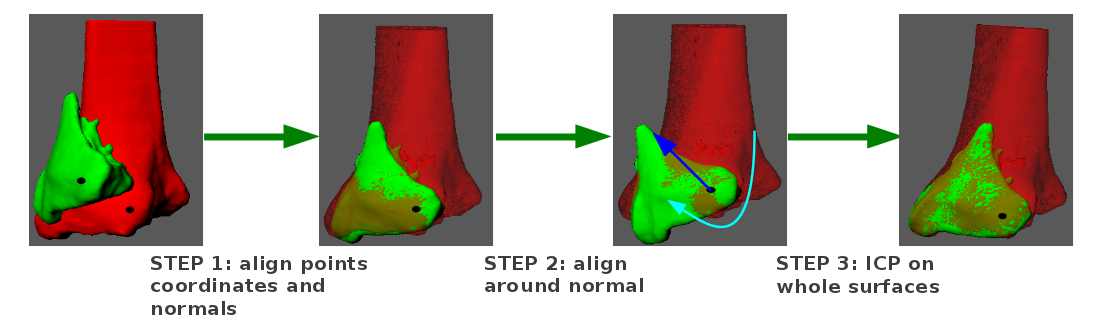}
\par\end{centering}
\caption[An example of alignment using the surface alignment engine]{\label{fig:An-example-of-alignment}An example of alignment using
the surface alignment engine. The red bone is the intact template,
the green bone is one of the fracture fragment. The two black points
on both surfaces are matched surface points.}
\end{figure}

Figure (\ref{fig:An-example-of-alignment}) illustrates how the alignment
process uses a hypothesized surface correspondence to align a fragment
to the intact template. For a given fragment, each hypothesized surface
correspondence is evaluated in the sequence given by the match score
from \S\eqref{subsec:Filtering-Feature-Matches} starting at the
highest score. Each evaluation aligns a fragment to the intact template
using the three alignment steps specified in the three previous section.
However, if the algorithm goes through all three steps for every match,
the reconstruction process is very time-consuming. To reduce computation
each alignment step includes predefined conditions to determine whether
the system should continue to evaluate the match or discard the match
and try a new hypothesized match. The local alignment error is used
to control the process. Because the computational cost increases in
each step, the alignment error threshold values are smaller (stricter)
for each step. In this paper, the threshold values are $4\gamma_{intact}$
for step one, $2\gamma_{intact}$ for step two, and $\gamma_{intact}$
for step three. Finally, when one match is accepted by all three steps
the output position from step three is considered as the final alignment
for the fragment.

\subsubsection{\label{subsec:Computational-Consideration}System Enhancements}

Several software enhancement tools are introduced to help improve
the quality of the 3D puzzle-solved solutions and to help reduce the
time necessary to compute these solutions. These enhancement tools
include: (1) surface sampling, (2) using occupied regions, (3) Mean
Curvature Histogram Biased Search, and (4) global fragment alignment
optimization. The following sections (\ref{subsec:Surface-Sampling},
\ref{subsec:Occupied-region}, \ref{subsec:Curvature-histogram},
\ref{subsec:Jiggling-Fragment}) discuss each tool in detail. 

\subsubsection{\label{subsec:Surface-Sampling}Surface Sampling}

The intact template and the fragment surfaces often consist\sout{s}
of a large number of surface points. Computing spin images for every
point on these surfaces is a time-consuming task. In order to improve
the speed of the puzzle-solving algorithm, uniform sub-sampling is
applied on both the intact template and the fragment surfaces. Surface
points are randomly selected on the fragment with a constraint that
the distance between any two sampled surface points are greater than
a predefined sampling distance $\triangle s$. The larger $\triangle s$
value computes fewer sampled points on the fragment surface and smaller
value computes more sampled points on the fragment surface. Spin images
are only computed for those sampled surface points on the intact template
and the fragment surfaces. In this paper, the sub-uniform sampling
distance for each fragment is set to $\triangle s=1.5\gamma_{intact}$,
and $\gamma_{intact}$ is the average edge length of edges on the
intact template.

\subsubsection{\label{subsec:Occupied-region}Using Occupied Regions}

The concept of occupied regions allows for significant performance
improvements by further reducing the spin images computed on the intact
template. Since part of the intact template surface has already been
aligned to the base fragment, these points should be excluded from
the matching. We mark these surface points on the intact template
surface as ``occupied'', and spin images are only computed for surface
points inside the ``unoccupied'' regions of the intact template.
This significantly reduces the search space for the matching process.
Occupied points on the intact template are flagged using the following
condition: the intact surface point, $\mathbf{v_{\textrm{j}}^{\textrm{t}}}$,
is marked as ``occupied'' if its distance to the closest aligned
fragment surface is less than $2\gamma_{intact}$. 

\subsubsection{\label{subsec:Curvature-histogram}Mean Curvature Histogram Biased
Search}

To reduce the number of spin images computed for both the intact template
and fragment surfaces, an approach called mean curvature histogram
biased search is used. Here we bias the search to emphasize regions
having discriminative shape by selectively computing descriptors in
locations where the mean surface curvature is large, i.e., highly
curved surface locations. The goal is to focus on calculating surface
descriptors that are more easily matched and have fewer candidate
correspondences. Using these locations allows computation to focus
on regions that carry more information regarding the unknown puzzle
solution. Towards this end, each descriptor is attributed with an
estimate of the surface mean curvature at the associated surface point.
We then randomly select points uniformly from the distribution of
observed curvatures which drastically reduces the number of spin images
computed. Further, when forming pairwise hypotheses we only create
hypotheses for points pairs whose mean surface curvature are similar.
This approach greatly reduces the computational cost of the puzzle
solving algorithm and improves performance of the reconstruction without
an observed sacrifice in solution accuracy.

\subsubsection{\label{subsec:Jiggling-Fragment}Global Optimization of Fragment
Alignment}

Global alignment seeks to simultaneously optimize all unknown transformation
parameters of the puzzle-solved fragments. This serves to correct
minor discrepancies in the position of misaligned fragments by fitting
together adjacent fragments onto the template surface in the final
reconstruction results. This is accomplished by perturbing fragment
transformation values away from the current puzzle solution and performing
ICP simultaneously using the global alignment algorithm of \citep{Pullia}.
This approach seeks to simultaneously equalized the surface alignment
error equally across all fragments and enforce the restriction that
fragment-and-template surface correspondences are unique, i.e., each
template surface point can only correspond with points from a single
fragment (as described in \ref{subsec:Occupied-region}). Since the
perturbations are small, this procedure serves to make small adjustments
each fragments final surface position on the intact template and minimize
fragment-to-fragment surface overlap. This process can improve the
pose of slightly misaligned fragments in the final puzzle solved solution.

\subsection{\label{subsec:Post-Reconstruction-Analysis}Post-Reconstruction Analysis}

The post-reconstruction analysis is the final output of the system.
The analysis tools are integrated into the system and allow users
to analyze the reconstruction result and help users better understand
the fracture case. The analysis consists of two components: (a) analysis
of the geometric accuracy of the aligned fragments and (b) analysis
of the severity of the bone fracture. The first component provides
a table containing alignment information, such as sampled, matched,
and unmatched points, and a histogram of alignment error for each
fragment. The second component is a fracture severity report which
contains quantitative values for several key factors for each fragment
which are known to be indicators of fracture severity, such as fracture
surface area. Although visual assessment of 3D reconstruction result
is valuable to users, these analysis tools provide quantitative information
which can help users objectively interpret the fracture case.

Analysis of the geometric accuracy of the reconstruction is accomplished
by measuring the point-to-plane distance from each vertex of the fragment
surface to the plane of the closest triangle on the intact template.
These distances are averaged to compute alignment error for each fragment.
The post-reconstruction analysis report includes a table of alignment
information and a table of histograms that show the distribution of
alignment errors for each fragment. 

\begin{figure}[H]
\begin{centering}
\subfloat[]{\includegraphics[width=0.6\columnwidth,height=0.25\textheight,keepaspectratio]{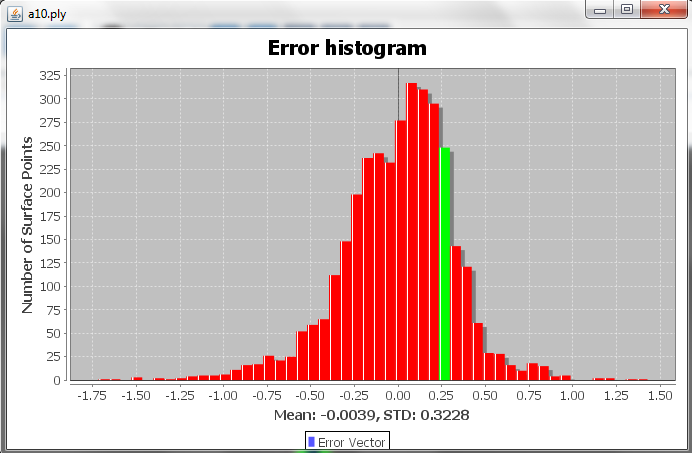}

}\subfloat[]{\includegraphics[width=0.4\columnwidth,height=0.25\textheight,keepaspectratio]{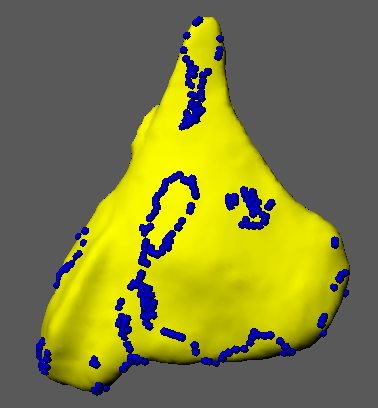}

}
\par\end{centering}
\caption[~Interactions on the histogram of alignment error]{\label{fig:histogramexample-1}(a) shows a histogram of alignment
errors where the bin of vertices having an error of approximately
0.25 $mm$ has been selected (green). (b) shows a visualization of
the spatial distribution surface points (blue spheres) which have
these errors on the fragment surface. }
\end{figure}

Our system provides histograms of the alignment error, providing users
with more detailed statistical information that relate to the quality
of the fragment alignment. Moreover, the system makes it possible
to visualize the location and magnitude of the alignment errors as
a spatial distribution on the fragment surface. Interactions are available
from the histogram plot which allow the user to select a range of
error values. Once selected, the errors associated with these values
are visualized spatially across the surface of the fragment as shown
in figure \ref{fig:histogramexample-1}. These tools are valuable
for understanding the quality of the geometric reconstruction results
and understanding the complex geometric inter-dependencies of a highly-fragmented
bone fracture. 

Severity assessment for bone fractures is heavily influenced by a
number of related key factors. One contribution of this paper is to
provide quantitative values for some of these key factors which are
heretofore unavailable in any other fracture analysis software. Figure
\ref{fig:fractureseveritytable} shows an example of the severity
assessment report which includes computed values of key factors for
each fragment in case one. The following list describes the values
provided in the severity assessment report in detail:
\begin{figure}
\begin{centering}
\includegraphics[width=0.95\columnwidth]{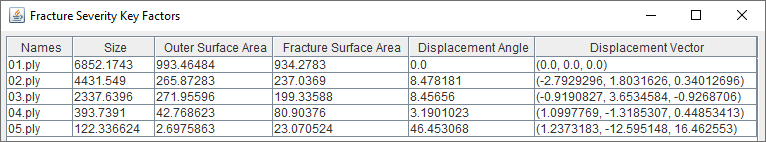}
\par\end{centering}
\caption[~Fracture severity key factors]{\label{fig:fractureseveritytable}A severity assessment report is
shown which includes values for several quantities that are known
to be key factors in determining fracture severity.}
\end{figure}

\begin{enumerate}
\item \textbf{Fragment Volume: }This is 3D volume of the region enclosed
by the fragment surface (shown as ``Size'' in the severity report
table). This information is useful in determining if the fragment
is structurally stable and useful in clinical reconstruction, i.e.,
can it be used for fixation, or is too small to be used in reconstruction.
\item \textbf{Fragment Fracture Surface Area}: The area of the fragment
fracture surface is a major factor in determining fracture severity
as shown in \citep{Beardsley2005}. The area of the fracture surface
generated is directly related to the energy that the fractured bone
absorbed during the fracture event. 
\item \textbf{Fragment Displacement:} The fragment displacement is the 3D
translation vector that moves the centroid of the fragment, which
indicates how far a fragment has moved or dispersed during the fracture
event.
\item \textbf{Angular Dislocation: }The angular dislocation of the fragment
is the angle between the principal axis of the bone fragment in its
original position and the principal axis of the bone fragment in its
reconstructed position. 
\end{enumerate}
These quantities are closely linked to fracture severity and the displayed
values may allow users to more accurately and objectively estimate
fracture severity. Computation of fracture surface areas from 2D or
3D CT images are difficult and the results are often unreliable. Here,
the total fracture surface area can be calculated more accurately
after the 3D fragment surfaces are segmented and anatomically classified.
Physicians have no way to quantitatively estimate fragment displacement
and angular dislocation from image data. Current approaches rely upon
the physician's visual assessment and experience. For high energy
fracture cases, accurately assessment of these values are difficult.
Since the proposed system has unique capabilities to compute the original
and reconstructed position for each bone fragment, the key factors
shown are inaccessible from any other source. Values uniquely available
from the proposed software are fragment displacement, angular dislocation,
and the surface area for the fragment outer surface and fracture surfaces.

\section{Results\label{chap:Results}}

The bone reconstruction system was used to reconstruct six clinical
fracture cases which range from low energy fracture events such as
1.5 foot fall, to high energy fracture events such as a 50 mph car
accident. The patient data, injury cause, and the Orthopedic Trauma
Association (OTA) classification for each case is shown in Table \ref{tab:Patient-demographic-and}.
Since these are real clinical cases, the patient names have been removed
to protect their privacy. All of the six cases were assigned a numerical
severity score ranging from (1-100) by three well-trained surgeons
based on their personal experience and subjective inference shown
in column $C_{1},$$C_{2}$, and $C_{3}$ of Table \ref{tab:Patient-demographic-and}.
Of particular note is the high variance across clinicians of these
classifications, consistent with the findings of \citep{Humphrey2005},
highlighting the need for objective measures of fracture severity.
\begin{table*}
\noindent \begin{centering}
\begin{tabular}{|c|c|c|c|c|c|c|c|c|}
\hline 
Case \# & Sex & Age & AO/OTA classification & Injury mechanism & C1 & C2 & C3 & Avg\tabularnewline
\hline 
\hline 
1 & F & 38 & C32 & MVA (50 mph) & 60 & 55 & 60 & 58\tabularnewline
\hline 
2 & M & 21 & B13 & Fall (30 ft) & 50 & 60 & 58 & 56\tabularnewline
\hline 
3 & F & 42 & C21 & MVA (30 mph) & 62 & 80 & 79 & 74\tabularnewline
\hline 
4 & M & 20 & C13 & ATV & 6 & 15 & 32 & 18\tabularnewline
\hline 
5 & M & 24 & C23 & Fall (12 ft) & 55 & 57 & 62 & 59\tabularnewline
\hline 
6 & M & 34 & C11 & Fall (18 ft) & 70 & 65 & 77 & 71\tabularnewline
\hline 
\end{tabular}
\par\end{centering}
\caption{\label{tab:Patient-demographic-and}Patient data, injury cause, OTA
classification, and severity scores by three surgeons for each case.
The higher severity scores indicate higher fracture severity.}
\end{table*}

The section \ref{sec:Case-Notes} shows the additional reconstruction
results for each case. For all six cases, the fragment and intact
surfaces were segmented, partitioned and classified using the methods
described in \ref{subsec:Segmenting-Fracture-CT}, \ref{subsec:Partitioning-Subsurfaces}
and \ref{subsec:Appearance-Based-3D-Surface} to provide the data
necessary to puzzle-solve each fracture case.
\begin{figure}
\begin{centering}
\includegraphics[height=0.45\textheight]{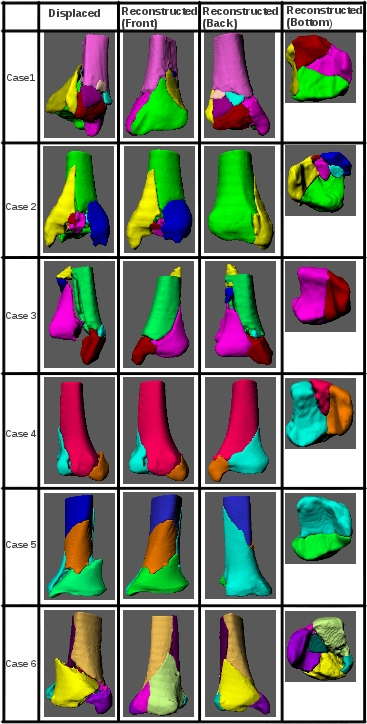}
\par\end{centering}
\caption[Reconstruction results of six clinical tibia plafond fractures]{\label{fig:Six-clinical-tibia}Six clinical tibia plafond fractures
are puzzle-solved. The original fractured positions for the fragments
are shown in the left column and three different views of reconstructed
fragments are shown in the remaining columns.}
\end{figure}

Figure \ref{fig:Six-clinical-tibia} shows the displaced positions
and three different views of reconstructed fragments for all six clinical
fractures. By visual assessment of the reconstruction results, all
cases were reconstructed successfully with the exception of case three.

\begin{table}
\centering{}%
\begin{tabular}{|c|>{\centering}p{0.2\columnwidth}|>{\centering}p{0.2\columnwidth}|>{\centering}p{0.2\columnwidth}|}
\hline 
Case ID & Completion time (sec) & Number of points on intact & Global Alignment Error (mm)\tabularnewline
\hline 
\hline 
1 & 140 & 24935 & 0.23\tabularnewline
\hline 
2 & 220 & 45529 & 0.27\tabularnewline
\hline 
3 & 272 & 50539 & 0.32\tabularnewline
\hline 
4 & 90 & 33630 & 0.34\tabularnewline
\hline 
5 & 430 & 68160 & 0.33\tabularnewline
\hline 
6 & 650 & 117549 & 0.27\tabularnewline
\hline 
\end{tabular}\caption{\label{tab:Puzzle-solving-performance.}Puzzle solving performance:
the time needed to run puzzle-solving algorithm for the fracture case
on a 2.4GHz, 4GB RAM laptop and the global alignment error.}
\end{table}
Column 3 of Table \ref{tab:Puzzle-solving-performance.} summarizes
the global alignment errors for all six cases. From the table, we
can see that global alignment error after the construction for all
six cases are relatively small (< 0.35mm), demonstrating higher accuracy
than competing methods \citep{Okada2009,Moghari2008,Fuernstahl2012}.
Table \ref{tab:Puzzle-solving-performance.} shows the time needed
to run the puzzle-solving algorithm. These reconstruction times were
recorded on a 2.4GHz, dual-core laptop with 4GB of RAM. The reconstruction
time recorded for each case includes time spent for computing spin
images, matching spin images, and aligning fragment surfaces. The
semi-automatic steps of the reconstruction: surface partitioning,
surface patch classification, and puzzle solving initialization are
not included here. From the table we can see that as the number of
points on the intact template increases so does the time required
for reconstruction. This is reasonable because more points result
in more hypotheses in the puzzle solving process. Indeed, the recorded
reconstruction time is influenced by several variables such as the
number of spin images computed from the intact template and the fragment
surfaces, number of fragments, number of iterations of the ICP algorithms,
etc. In \S\ref{sec:Performance-Improvement-Results} we discuss a
small modification made to boost performance. Table \ref{tab:total-summary}
in the Appendix records the results of each fragment processed by
the system.

\subsection{Notes on Each Case\label{sec:Case-Notes}}

Each of the six cases processed by our system supplies valuable insight
to the versatility of the reconstruction method. All cases except
Case 3 (which will be explained below) were deemed successful due
to good visual geometric agreement and via the value of the global
alignment error. Case 1 contained ten unique fragments to place, yet
still achieved a global error of 0.23mm. This demonstrates the efficacy
of the proposed method even in complex fracture cases. Case 2's fragment
A4 underwent a displacement magnitude of almost five centimeters (48.98mm),
which is substantially large compared to the total size of the tibia's
epiphysis. Despite this, Case 2's reconstruction was successful, demonstrating
the resilience of the proposed reconstruction approach to local minima.
Case 3 was the only unsuccessful reconstruction of all test cases.
Its fragment A2 clipped through the base fragment (A1) despite the
anatomical impossibility. This may be attributed to the low resolution
of Case 3's CT DICOMs. It had the lowest resolution at 0.683mm per
pixel which would increase the likelihood of false positive matches
in the generated spin images. Case 4 and 5 consist of a small number
of fragments (3 and 4 respectively). This resulted in higher points
per fragment to match, increasing the computational cost of the puzzle
solution. However, reconstruction for this case was successful and
demonstrates the system's capacity for processing large fragment surface
matches. Case 6 was unique in its low percentages of outer surface
area on each fragment. The reconstruction was still successful despite
the smaller amount of points to match per fragment.
\begin{table*}
\begin{tabular}{|>{\centering}p{0.06\textwidth}|>{\centering}p{0.1\textwidth}|>{\centering}p{0.1\textwidth}|>{\centering}p{0.12\textwidth}|>{\centering}p{0.12\textwidth}|>{\centering}p{0.08\textwidth}|>{\centering}p{0.06\textwidth}|>{\centering}p{0.15\textwidth}|}
\hline 
Case & $t_{tl}$ w/o accel. (min) & $t_{tt}$ w/ accel. (min) & Avg match\sout{ }time\sout{ }w/o accel.(sec) & Avg matching w/ accel.(sec) & $n_{i}$ 

w/o accel. & $n_{i}$ 

w/ accel. & \# Points On Template\tabularnewline
\hline 
\hline 
1 & 3 & 1.5 & 11 & 0.3 & 3750 & 1125 & 24935\tabularnewline
\hline 
2 & 6 & 4.5 & 15 & 0.8 & 5065 & 802 & 45529\tabularnewline
\hline 
3 & 8 & 5 & 16 & 1.2 & 5320 & 1913 & 50539\tabularnewline
\hline 
4 & 2.5 & 1 & 10 & 0.2 & 2509 & 897 & 33630\tabularnewline
\hline 
5 & 20 & 6 & 34 & 2.4 & 8890 & 4135 & 68160\tabularnewline
\hline 
6 & 31 & 16 & 52 & 3.5 & 16829 & 3120 & 117549\tabularnewline
\hline 
\end{tabular}\caption{\label{tab:The-table-ofperformance}Computational performance observed
for each case using the system enhancements discussed in \textsection
\ref{subsec:Curvature-histogram}.}
\end{table*}

\subsection{\label{sec:Performance-Improvement-Results}Performance Improvements}

As mentioned in \S\ref{subsec:Curvature-histogram}, the mean curvature
histogram biased search algorithm significantly improves the speed
of the automatic puzzle-solving algorithm by reducing the number of
spin images computed for both the intact template surface and the
fragment surfaces. The puzzle-solving algorithm is a complex process
which consists of many steps, and the time spent for the reconstruction
is affected by intermediate steps such as computing the spin images
for the intact template and the fragment surfaces, matching spin images,
and aligning fragment surfaces to the intact template. In order to
better understand the improvements, the following equation (\ref{eq:reconstructiontime})
details the time spent for reconstruction for each step.
\begin{equation}
t_{tl}=t_{o}+n_{i}t_{s}+\left(t_{s}n_{f}+t_{m}+t_{f}+t_{a}\right)M_{f}\label{eq:reconstructiontime}
\end{equation}

In this equation, $t_{o}$ denotes the time spent for processing the
intact template before computing the spin images on the intact template
such as surface sampling, computing occupied regions and computing
the mean curvatures for the intact template. The term $n_{i}t_{s}$
denotes the time spent to identifying feature points on the intact
template and compute their spin images. The term $t_{s}n_{f}$ denotes
the time spent computing descriptors on the fragment surfaces. The
term $t_{m}$ denotes time spent generating hypothesized surface correspondences
which is affected by $n_{f}$ and $n_{i}$. The term $t_{f}$ denotes
the time spent removing false matches. The term $t_{a}$ denotes time
spent aligning the fragment surface to the intact template which is
affected by number of hypothesized correspondences being tested. The
term ${\color{blue}K}$ denotes the number of fragments in the fracture
case. The major factors that impact the total reconstruction time
are $n_{f}$, the number of spin images computed on the fragment surface,
and $n_{i}$, the number spin images computed on the intact template.
The mean curvature histogram approach in \S\ref{subsec:Curvature-histogram}
reduces the total reconstruction time by reducing both $n_{i}$ and
$n_{f}$ significantly Table \ref{tab:The-table-ofperformance} shows
the quantitative values for total reconstruction time, $t_{tl}$,
average time spent for matching spin images and filtering matches,
$t_{m}+t_{f}$, and $n_{i}$ for each case. Note that acceleration
provides roughly an order of magnitude decrease on computation time
for matching and reduces the total reconstruction time, including
user interaction, by about 50\%. Experimental reconstructions were
conducted using a laptop computer with 2.4GHz dual core CPU with 4GB
memory.

Our accelerated reconstruction times are significantly shorter than
those reported in similar systems. Work in \citep{Fuernstahl2012}
reported an average total reconstruction time of 83 minutes which
is approximately 5 times longer than our worst-case clinical reconstruction
(16 mins) and more than 10 times longer than the average time it requires
for reconstruction across our 6 cases (\textasciitilde{}5 mins 45
secs). Unfortunately the performance in time of the methods described
in \citep{Okada2009,Moghari2008} are not reported.

\section{\label{cha:Conclusion-And-Proposed}Conclusion}

The proposed system is capable of virtually reconstructing broken
bone fragments for complex bone fracture cases, which is currently
an unsolved problem in automatic puzzle-solving algorithms and difficult
to achieve using manual methods. The bone reconstruction system designed
in this paper enables users to understand fracture cases from both
2D (CT image) and 3D (fragment surface) imagery. The system represents
a unique combination of state-of-the-art 2D/3D image processing and
surface processing algorithms. The software is a comprehensive reconstruction
tool that guides users from the first step, i.e., segmenting raw CT
image data, to the last step, i.e., generating quantitative, critical
information about the fracture's severity. Finally, 3D visualization
of fragment surfaces can provide important information for surgical
treatment, especially for articular fractures which often have a poor
prognosis.

The system demonstrates the efficacy of spin image reconstruction
in bone fractures from various fracture events with a wide variety
of traits. Reconstruction was successful in bone fractures with both
many and few fragments (Cases 1, 4, and 5), fractures where fragments
have undergone large displacements (Case 2), and fractures where little
intact outer surface area remains (Case 6). The reconstruction only
suffers in cases where the source CT images were generated at low
resolutions (greater than 0.5mm per pixel, Case 3). In each case,
global alignment errors less than 0.35mm were achieved.

With the detailed fracture analysis offered by this tool, it could
potentially improve surgical treatment, as previously explored in
\citep{Thomas2010,Thomas2011}. The system represents a significant
step towards assisting physicians in classifying fracture severity
which is especially difficult in high-energy fracture cases. Use of
quantitative metrics in conjunction with visual assessment promises
to reduce variability in fracture severity classification and may
serve to build consensus in these difficult cases. The puzzle-solving
algorithms and their integration as a software system represent significant
advancements toward improving the treatment of comminuted tibial plafond
fractures and it is entirely possible this technology can be applied
to other problematic limb fracture cases. While not directly discussed
in this article we also note that the computational 3D puzzle solving
framework provides a heretofore unavailable patient-specific blueprint
for fracture reconstruction planning. Having a suitable blueprint
for restoring the original anatomy, it becomes possible for the surgeon
to pre-operatively explore less extensive surgical approaches and
to attempt new minimally invasive surgical approaches.

\section*{Appendix A. Summary of per-fragment results for reconstruction\textcolor{blue}{\label{sec:Appendix-A}}}

Table \ref{tab:total-summary} records the results of each fragment
processed by the system. Displacement and displaced angle demonstrate
how far each fragment moved from its reconstructed position in the
fracture event. The fracture surface area provides insight to the
energy of the event. The percentage of outer surface area to total
surface area provides insight to the difficulty of finding matches
to the intact bone template.
\begin{sidewaystable*}
\noindent \centering{}%
\begin{tabular}{|c|c|c|c|c|c|c|}
\cline{2-7} 
\multicolumn{1}{c|}{} & Fragment & Displacement (mm) & Displaced Angle (deg) & Outer Surface Area (mm\textsuperscript{2}) & Fracture Surface Area (mm\textsuperscript{2}) & \% Outer to Total Area\tabularnewline
\hline 
\multirow{10}{*}{\begin{turn}{90}
Case 1
\end{turn}} & A1 & 0 & 0 & 274.5 & 446.4 & 45.62\tabularnewline
\cline{2-7} 
 & A2 & 7.201 & 20.40 & 91.96 & 115.5 & 44.33\tabularnewline
\cline{2-7} 
 & A3 & 2.256 & 6.541 & 4.037 & 4.382 & 47.95\tabularnewline
\cline{2-7} 
 & A4 & 13.07 & 29.19 & 126.8 & 223.7 & 26.18\tabularnewline
\cline{2-7} 
 & A5 & 6.324 & 17.60 & 43.86 & 48.68 & 47.40\tabularnewline
\cline{2-7} 
 & A6 & 8.566 & 15.29 & 11.74 & 11.21 & 51.16\tabularnewline
\cline{2-7} 
 & A7 & 13.30 & 28.38 & 133.7 & 236.6 & 36.11\tabularnewline
\cline{2-7} 
 & A8 & 2.948 & 5.765 & 12.63 & 15.70 & 44.58\tabularnewline
\cline{2-7} 
 & A9 & 12.98 & 26.25 & 74.39 & 164.2 & 31.18\tabularnewline
\cline{2-7} 
 & A10 & 16.89 & 37.34 & 29.58 & 29.35 & 50.20\tabularnewline
\hline 
\multirow{6}{*}{\begin{turn}{90}
Case 2
\end{turn}} & A1 & 0 & 0 & 949.8 & 993.9 & 48.87\tabularnewline
\cline{2-7} 
 & A2 & 6.838 & 12.19 & 389.8 & 338.9 & 53.49\tabularnewline
\cline{2-7} 
 & A3 & 7.481 & 20.03 & 218.8 & 159.3 & 57.87\tabularnewline
\cline{2-7} 
 & A4 & 48.98 & 85.92 & 17.10 & 108.1 & 13.66\tabularnewline
\cline{2-7} 
 & A5 & 29.78 & 178.4 & 44.0 & 29.62 & 59.78\tabularnewline
\cline{2-7} 
 & A6 & 21.59 & 44.29 & 7.449 & 15.62 & 32.29\tabularnewline
\hline 
\multirow{6}{*}{\begin{turn}{90}
Case 3
\end{turn}} & A1 & 0 & 0 & 2331 & 1487 & 61.05\tabularnewline
\cline{2-7} 
 & A2 & 38.12 & 62.87 & 216.2 & 117.4 & 54.93\tabularnewline
\cline{2-7} 
 & A3 & 35.15 & 36.51 & 91.63 & 110.9 & 45.24\tabularnewline
\cline{2-7} 
 & A4 & 22.59 & 10.79 & 1848 & 1143 & 61.79\tabularnewline
\cline{2-7} 
 & A5 & 29.15 & 25.26 & 55.65 & 62.12 & 47.25\tabularnewline
\cline{2-7} 
 & A6 & 83.56 & 47.64 & 574.9 & 907.8 & 38.77\tabularnewline
\hline 
\multirow{3}{*}{\begin{turn}{90}
Case 4
\end{turn}} & A1 & 0 & 0 & 1924 & 1754 & 52.31\tabularnewline
\cline{2-7} 
 & A2 & 3.545 & 4.931 & 680.8 & 859.0 & 44.21\tabularnewline
\cline{2-7} 
 & A3 & 3.734 & 12.25 & 380.6 & 464.2 & 45.05\tabularnewline
\hline 
\multirow{4}{*}{\begin{turn}{90}
Case 5
\end{turn}} & A1 & 0 & 0 & 463.3 & 438.9 & 51.35\tabularnewline
\cline{2-7} 
 & A2 & 2.987 & 5.175 & 851.9 & 755.6 & 52.99\tabularnewline
\cline{2-7} 
 & A3 & 1.631 & 5.101 & 303.3 & 332.8 & 47.68\tabularnewline
\cline{2-7} 
 & A4 & 15.79 & 23.12 & 330.6 & 463.5 & 41.63\tabularnewline
\hline 
\multirow{8}{*}{\begin{turn}{90}
Case 6
\end{turn}} & A1 & 0 & 0 & 246.8 & 246.0 & 50.08\tabularnewline
\cline{2-7} 
 & A2 & 5.979 & 16.89 & 17.26 & 31.14 & 35.66\tabularnewline
\cline{2-7} 
 & A3 & 1.015 & 3.992 & 108.7 & 159.0 & 40.61\tabularnewline
\cline{2-7} 
 & A4 & 2.778 & 20.08 & 93.70 & 292.6 & 24.26\tabularnewline
\cline{2-7} 
 & A5 & 2.887 & 14.27 & 28.03 & 54.24 & 34.07\tabularnewline
\cline{2-7} 
 & A6 & 7.564 & 19.17 & 10.03 & 51.67 & 16.26\tabularnewline
\cline{2-7} 
 & A7 & 1.686 & 8.294 & 97.33 & 249.0 & 28.10\tabularnewline
\cline{2-7} 
 & A8 & 8.417 & 23.45 & 46.44 & 133.8 & 25.77\tabularnewline
\hline 
\end{tabular}\caption{\label{tab:total-summary}Summary of results for all fragments of
the six test cases.}
\end{sidewaystable*}

\begin{verse}
\noindent \bibliographystyle{elsarticle-harv}
\bibliography{2019_BonePuzzleSolvingAlgorithms,thesis_proposal}
\end{verse}

\end{document}